\documentclass[sigconf]{acmart}

\renewcommand\footnotetextcopyrightpermission[1]{} % removes footnote with conference information in first column
\usepackage{float}
\usepackage{amsmath}
\usepackage{subfigure}
\usepackage{multirow}
\usepackage{booktabs}
\usepackage{lscape}
\usepackage{balance}

\AtBeginDocument{%
  \providecommand\BibTeX{{%
    \normalfont B\kern-0.5em{\scshape i\kern-0.25em b}\kern-0.8em\TeX}}}

\acmConference[]{}{}{}
\begin{document}

%%
%% The "title" command has an optional parameter,
%% allowing the author to define a "short title" to be used in page headers.
\title{Automating Visual Blockage Classification of Culverts with Deep Learning}

%%
%% The "author" command and its associated commands are used to define
%% the authors and their affiliations.
%% Of note is the shared affiliation of the first two authors, and the
%% "authornote" and "authornotemark" commands
%% used to denote shared contribution to the research.

\author{Umair Iqbal}
\authornotemark[1]
\affiliation{%
  \institution{SMART, University of Wollongong}
 \city{Wollongong}
  \state{NSW}
  \country{Australia}
}
\email{ui010@uowmail.edu.au}

\author{Johan Barthelemy}
\affiliation{%
  \institution{SMART, University of Wollongong}
  \city{Wollongong}
 \state{NSW}
  \country{Australia}
}
\email{johan@uow.edu.au}

\author{Wanqing Li}
\affiliation{%
  \institution{SCIT, University of Wollongong}
 \city{Wollongong}
  \state{NSW}
  \country{Australia}
}
\email{wanqing@uow.edu.au}

\author{Pascal Perez}
\affiliation{%
  \institution{SMART, University of Wollongong}
  \city{Wollongong}
  \state{NSW}
  \country{Australia}
}
\email{pascal@uow.edu.au}

%% By default, the full list of authors will be used in the page
%% headers. Often, this list is too long, and will overlap
%% other information printed in the page headers. This command allows
%% the author to define a more concise list
%% of authors' names for this purpose.

\renewcommand{\shortauthors}{Iqbal et al.}

%%
%% The abstract is a short summary of the work to be presented in the
%% article.
\begin{abstract}
  Blockage of culverts by transported debris materials is reported as main contributor in originating urban flash floods. Conventional modelling approaches had no success in addressing the problem largely because of unavailability of peak floods hydraulic data and highly non-linear behaviour of debris at culvert. This article explores a new dimension to investigate the issue by proposing the use of Intelligent Video Analytic (IVA) algorithms for extracting blockage related information. Potential of using existing Convolutional Neural Network (CNN) algorithms (i.e., DarkNet53, DenseNet121, InceptionResNetV2, InceptionV3, MobileNet, ResNet50, VGG16, EfficientNetB3, NASNet) is investigated over a custom collected blockage dataset (i.e., Images of Culvert Openings and Blockage (ICOB)) to predict the blockage in a given image. Models were evaluated based on their performance on test dataset (i.e., accuracy, loss, precision, recall, F1-score, Jaccard-Index), Floating Point Operations Per Second (FLOPs) and response times to process a single test instance. From the results, NASNet was reported most efficient in classifying the blockage with the accuracy of 85\%; however, EfficientNetB3 was recommended for the hardware implementation because of its improved response time with accuracy comparable to NASNet (i.e., 83\%). False Negative (FN) instances, False Positive (FP) instances and CNN layers activation suggested that background noise and oversimplified labelling criteria were two contributing factors in degraded performance of existing CNN algorithms.  
\end{abstract}

%%
%% The code below is generated by the tool at http://dl.acm.org/ccs.cfm.
%% Please copy and paste the code instead of the example below.
%%



%%
%% Keywords. The author(s) should pick words that accurately describe
%% the work being presented. Separate the keywords with commas.

\maketitle

\section{Introduction}

Cross-drainage structures (e.g., culverts, bridges) are prone to blockage by debris and reported as one of the main causes of flash floods in urban areas \cite{french2015culvert, french2012non, blanc2013analysis, ARR_Report, roso2004prediction, wallerstein1996debris, iqbalcomputer}. The 1998 and 2011 floods in Wollongong, Australia \cite{french2015culvert, BarthelmessMechanism, rigby2002causes, van2001modelling, davis2001analysis} and the 2007 flood in Newcastle, Australia \cite{french2015culvert, wbm2008newcastle} are classical examples where blockage of cross drainage hydraulic structures caused the flash flooding. Project 11: Blockage of Hydraulic Structures \cite{ARR_Report} was initiated under the Australian Rainfall and Runoff (ARR) \cite{ball2016australian} framework to study the blockage behaviour and design considerations of hydraulic structures. Under this project, Wollongong City Council (WCC) proposed the guidelines to consider the hydraulic blockage in the hydraulic structures design process \cite{ARR_Report, french2018design, rigby2001impact, ollett2017australian, CondouitBlockagePolicy, french2012non}. However, because of the unavailability of relevant supporting data from peak flooding events, proposed guidelines were not adaptive and were based on the post flood visual assessments, which many researchers believe is not the correct representation of blockage during the peak flooding events \cite{french2015culvert, french2012non, french2018design}. The guidelines suggested that any culvert with an opening diagonal of 6m or more is not prone to blockage. However, this claim was only supported by post flood visual assessments and was not considered economically efficient to implement. 

Initially, blockage was defined as the percentage occlusion of hydraulic structure opening, however, many argued that hydraulic blockage and visual blockage are two separate concepts. Hydraulic blockage is more complex and has no established relationship with visual blockage. Hydraulic blockage is associated with the interaction of debris with culvert and corresponding effect on fluid dynamics around culvert, however, due to highly non-linear and uncertain behaviour of debris, it is difficult to model and predict the hydraulic blockage using conventional means. From management and maintenance perspective, making use of multi-dimensional information (i.e., visual blockage status, type of debris material, percentage of blocked openings) extracted using computer vision algorithms may prove helpful in making timely decisions as suggested in literature \cite{iqbalcomputer, arshad2019computer}. This paper attempts to address the problem from a different perspective and proposes the use of visual information extracted using automated analysis in better management of blockage at cross drainage hydraulic structures. 

This paper investigated the potential of CNN algorithms towards classifying culvert images as ``clear" or ``blocked". Existing CNN models (i.e., DarkNet53 \cite{redmon2018yolov3}, DenseNet121 \cite{huang2017densely}, InceptionResNetV2 \cite{szegedy2016inception}, InceptionV3 \cite{szegedy2016rethinking}, MobileNet \cite{howard2017mobilenets}, ResNet50 \cite{he2016deep}, VGG16 \cite{simonyan2014very}, EfficientNetB3 \cite{tan2019efficientnet}, NASNet \cite{zoph2018learning}) pre-trained over ImageNet were transfer-learned for the culvert blockage classification task and performance was compared based on the standard evaluation measures.

\begin{figure*}[ht]
    \centering
    \includegraphics[scale=0.98]{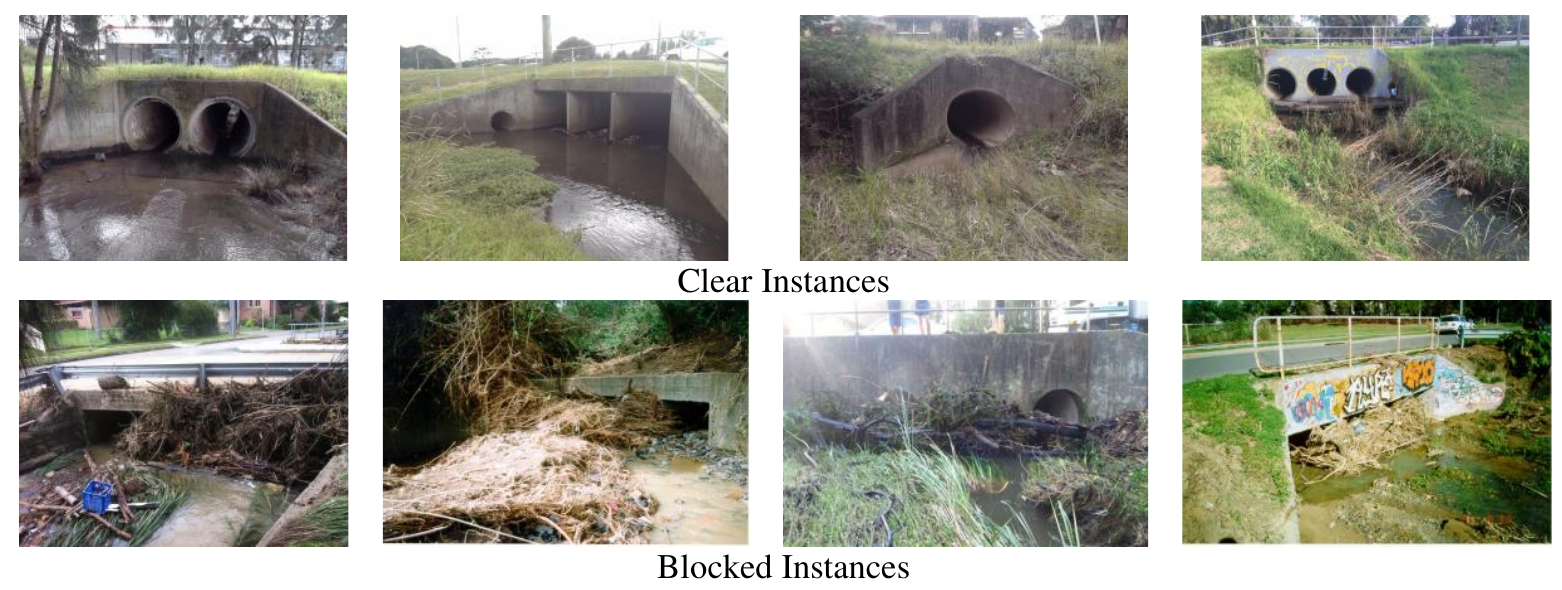}
    \caption{Sample Instances of Clear (First Row) and Blocked (Second Row) Culverts from ICOB.}
    \label{fig:Real_Samples}
\end{figure*}

\section{ICOB Dataset}

The dataset used for this investigation is referred as ``Images of Culvert Openings and Blockage (ICOB)" and consisted of real culverts images collected before and after the flooding events. Main sources of images included WCC historical records, online records and custom captured local culvert images. WCC records were scrutinized using a Microsoft ACCESS based application for filtering the culvert images with visible openings. Final dataset included 929 images of culverts including both blocked and clear. Dataset contained images with high level of variation from each other (intra-class variation) in terms of culvert types, blockage accumulation, presence of debris materials, illumination conditions, culvert view point variations, scale variations, resolution, and backgrounds. This high level of diversity within a relatively small dataset makes it a challenging dataset for visual analytic, even for a binary classification problem.

ICOB dataset was manually labelled for binary classification of a given image with culvert as ``clear" or ``blocked". A culvert being visually blocked or clear is not as simple and may require defining a detailed criteria in collaboration with flood management officers; however, for this article, simple occlusion based criteria was used. Following subjective annotation criteria was used for labelling ICOB. 
\begin{itemize}
\item If all of the culvert openings are visible, classify it as ``clear".
\item If any of the culvert opening is visually occluded by debris material or foreground object (e.g., debris control structure, vegetation, tree), classify it as ``blocked".
\end{itemize}

 In total, there were 487 images in ``clear" class and  442 images in ``blocked " class. Figure \ref{fig:Real_Samples} shows the sample instances from each class of ICOB.

\begin{table*}
    \caption{Classification Performance of Implemented CNN Models for Blockage Detection.}
    \label{tab:performance}
    \centering
    \begin{tabular}{cccccccc}
    \toprule
    ~ & \textbf{Test Accuracy} & \textbf{Test Loss} & \textbf{Precision Score} & \textbf{Recall Score} & \textbf{F1 Score} & \textbf{Jaccard-Index} & \textbf{FLOPs}\\
    \toprule
DarkNet53	& 0.71	& 1.22	& 0.72	& 0.71	& 0.71	& 0.55 & 14.2 G \\
DenseNet121	& 0.79	& 0.48	& 0.79	& 0.79	& 0.79	& 0.65 & 5.7 G\\
InceptionResNetV2	& 0.83	& 0.46	& 0.84	& 0.83	& 0.83	& 0.71 & 13.3 G\\
InceptionV3	& 0.81	& 0.50	& 0.81	& 0.81	& 0.81	& 0.68 & 5.69 G\\
MobileNet	& 0.79	& 0.50	& 0.80	& 0.80	& 0.79	& 0.66 & \textbf{1.15 G}\\
ResNet50	& 0.80	& 0.52	& 0.81	& 0.81	& 0.80	& 0.67 & 7.75 G\\
VGG16	& 0.75	& 0.56	& 0.75	& 0.75	& 0.75	& 0.59 & 30.7 G\\
EfficientNetB3	& 0.83	& \textbf{0.45}	& 0.83	& 0.83	& 0.83	& 0.7 & 1.97 G\\
NASNetLarge	& \textbf{0.85}	& 0.54	& \textbf{0.86}	& \textbf{0.85}	& \textbf{0.85}	& \textbf{0.74} & 47.8 G\\
\bottomrule
    \end{tabular}
\end{table*}

\begin{figure*}
\centering
\subfigure[DarkNet53]{\includegraphics[scale=0.35]{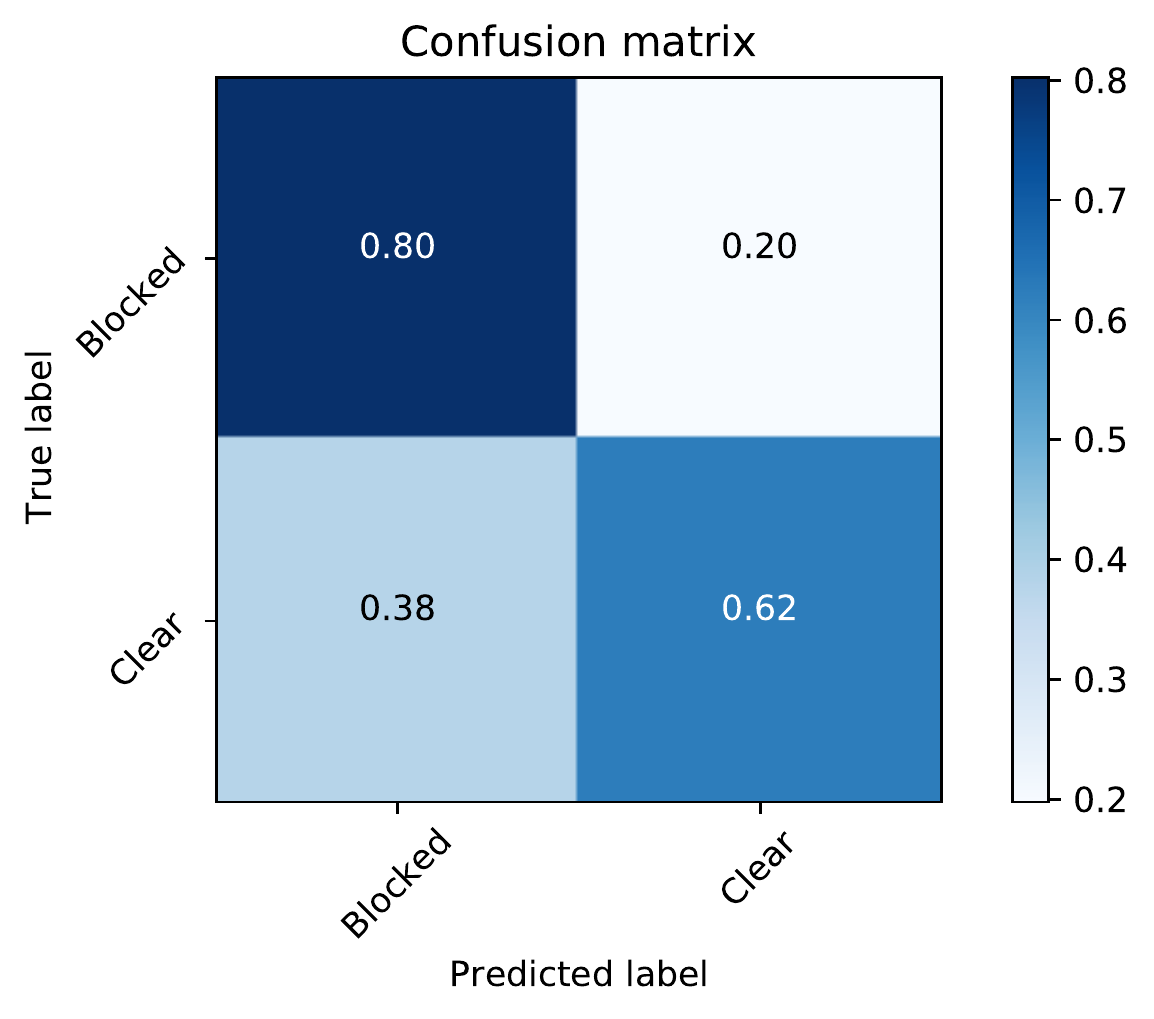}}
\subfigure[DenseNet121]{\includegraphics[scale=0.35]{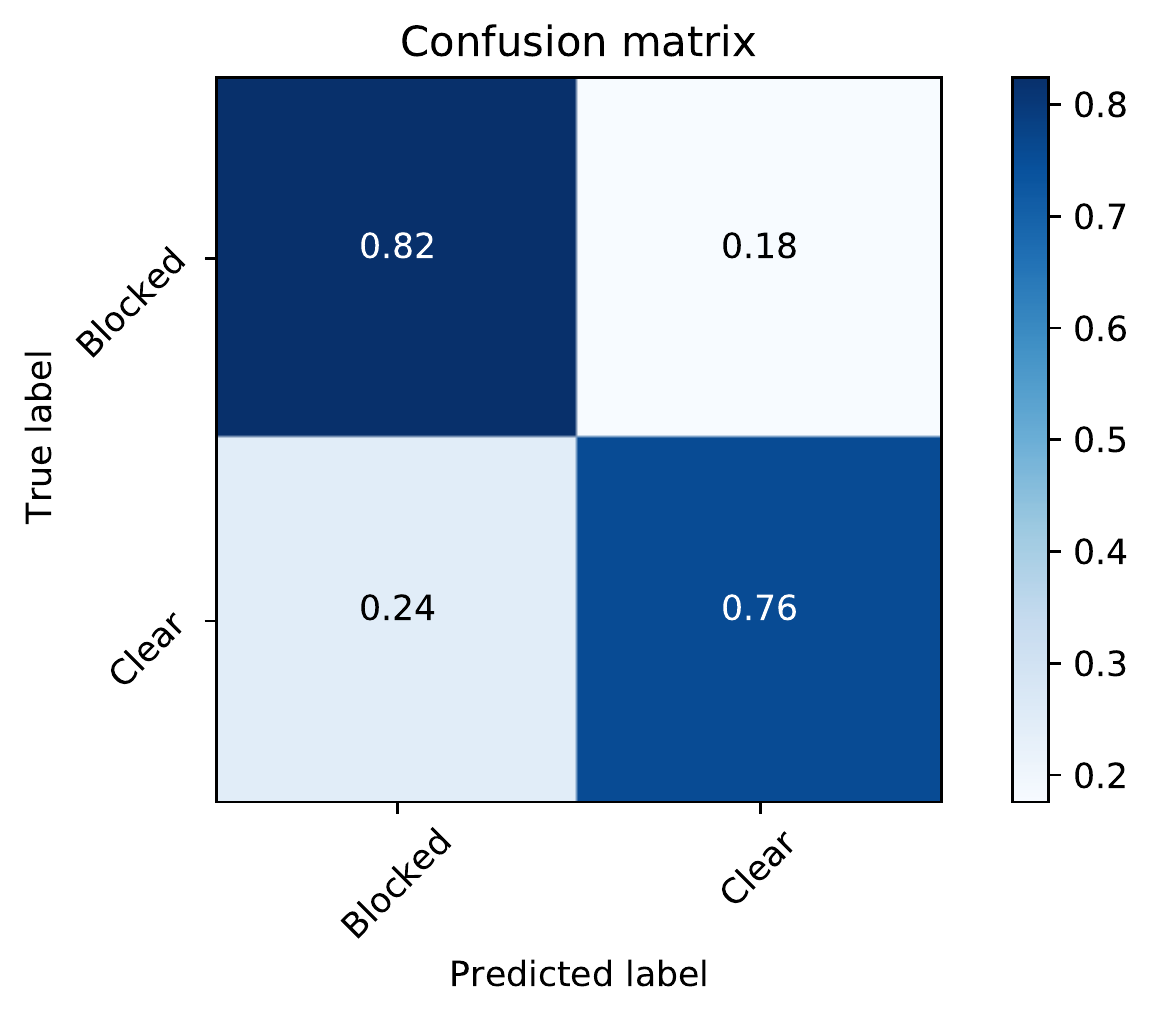}}
\subfigure[InceptionResNetV2]{\includegraphics[scale=0.35]{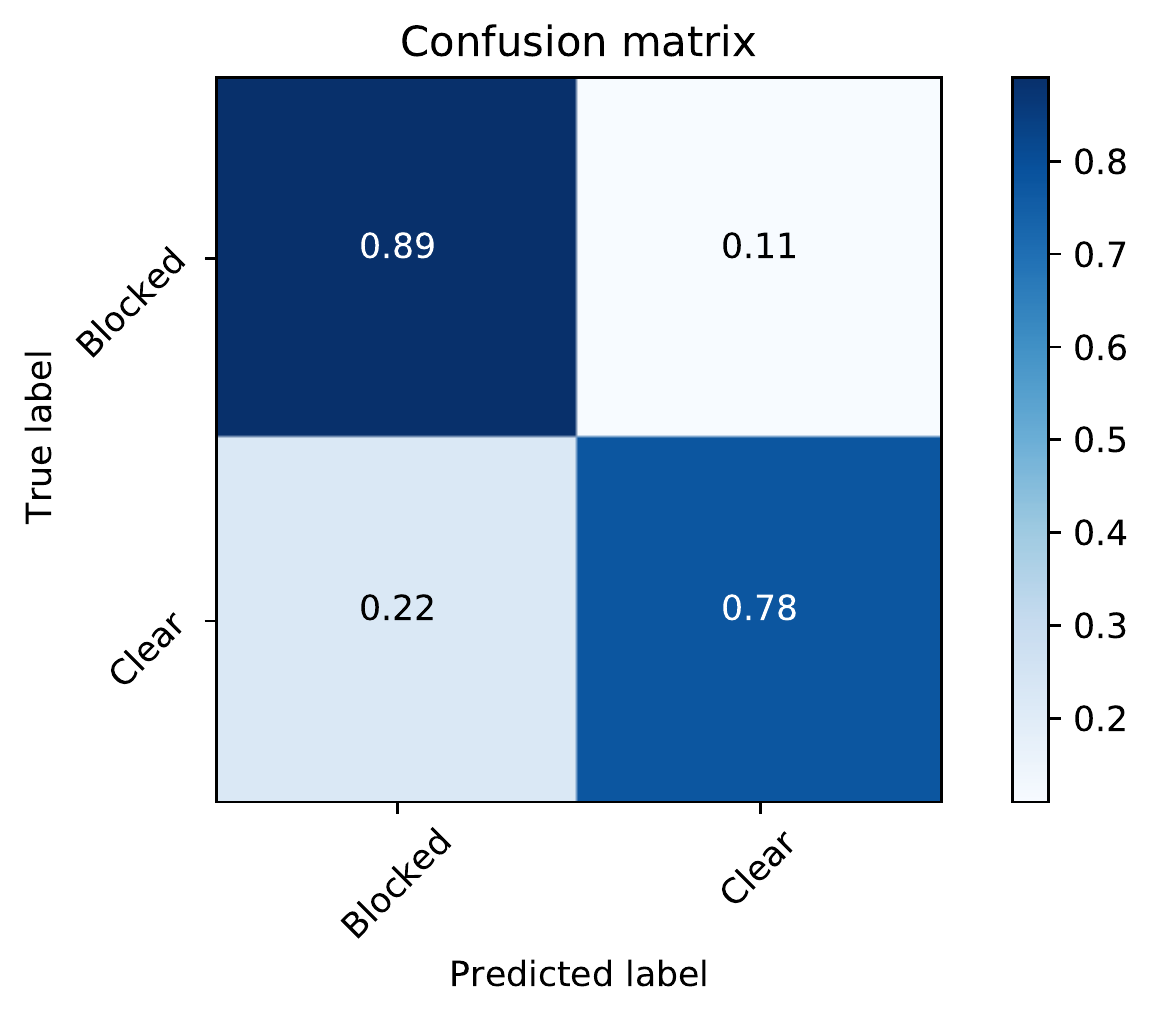}}
\subfigure[InceptionV3]{\includegraphics[scale=0.35]{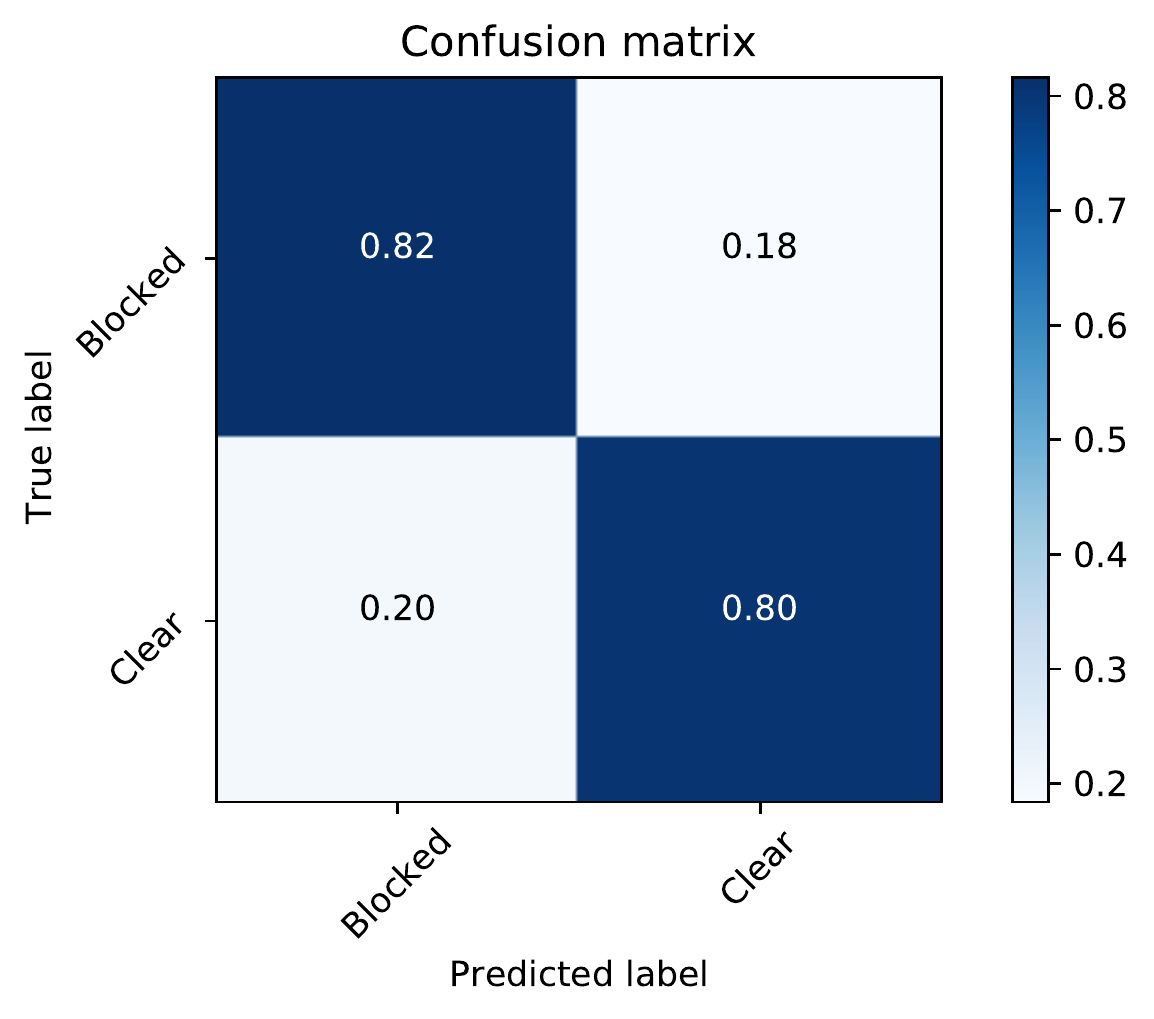}}\\
\subfigure[MobileNet]{\includegraphics[scale=0.35]{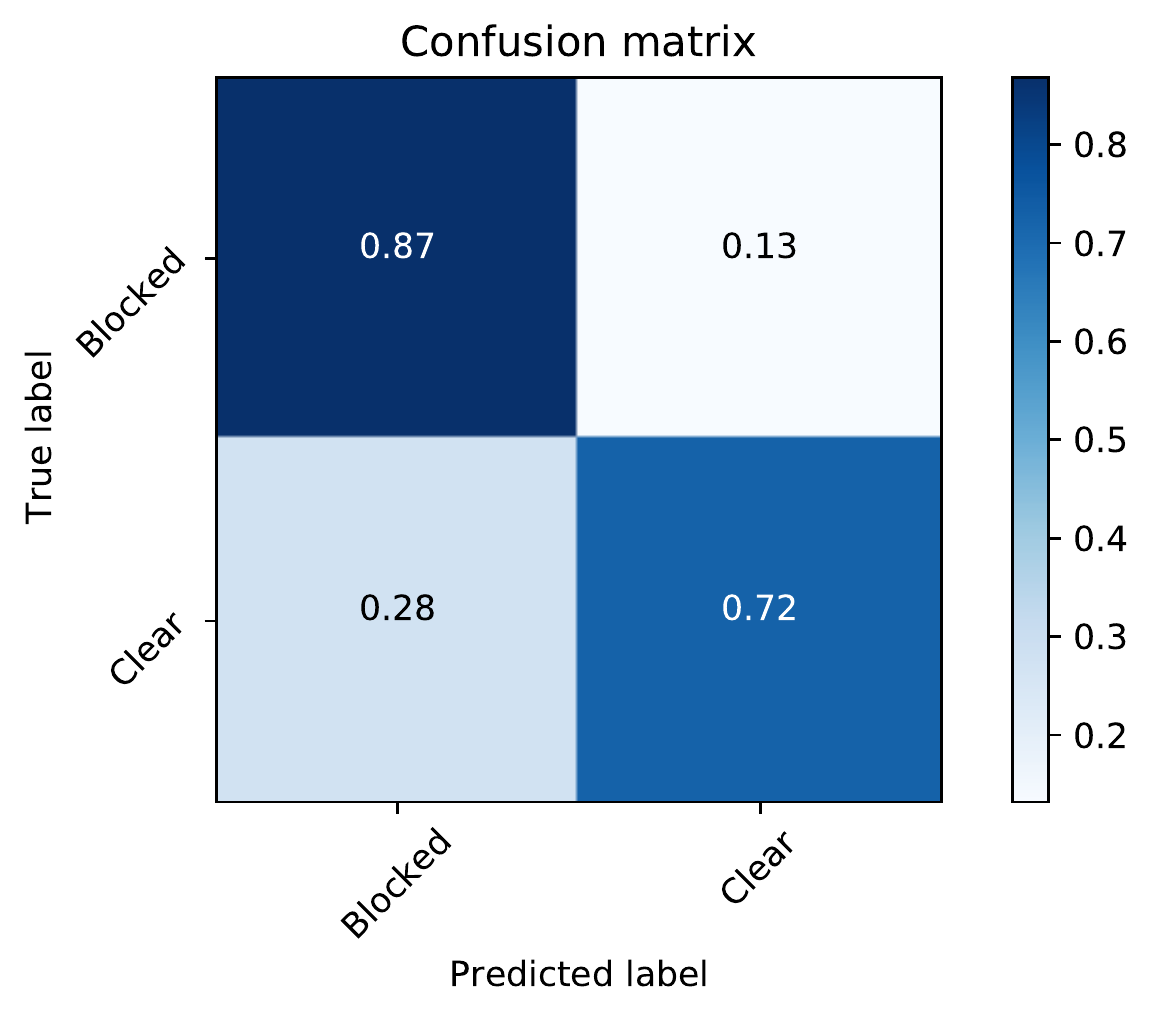}}
\subfigure[ResNet50]{\includegraphics[scale=0.35]{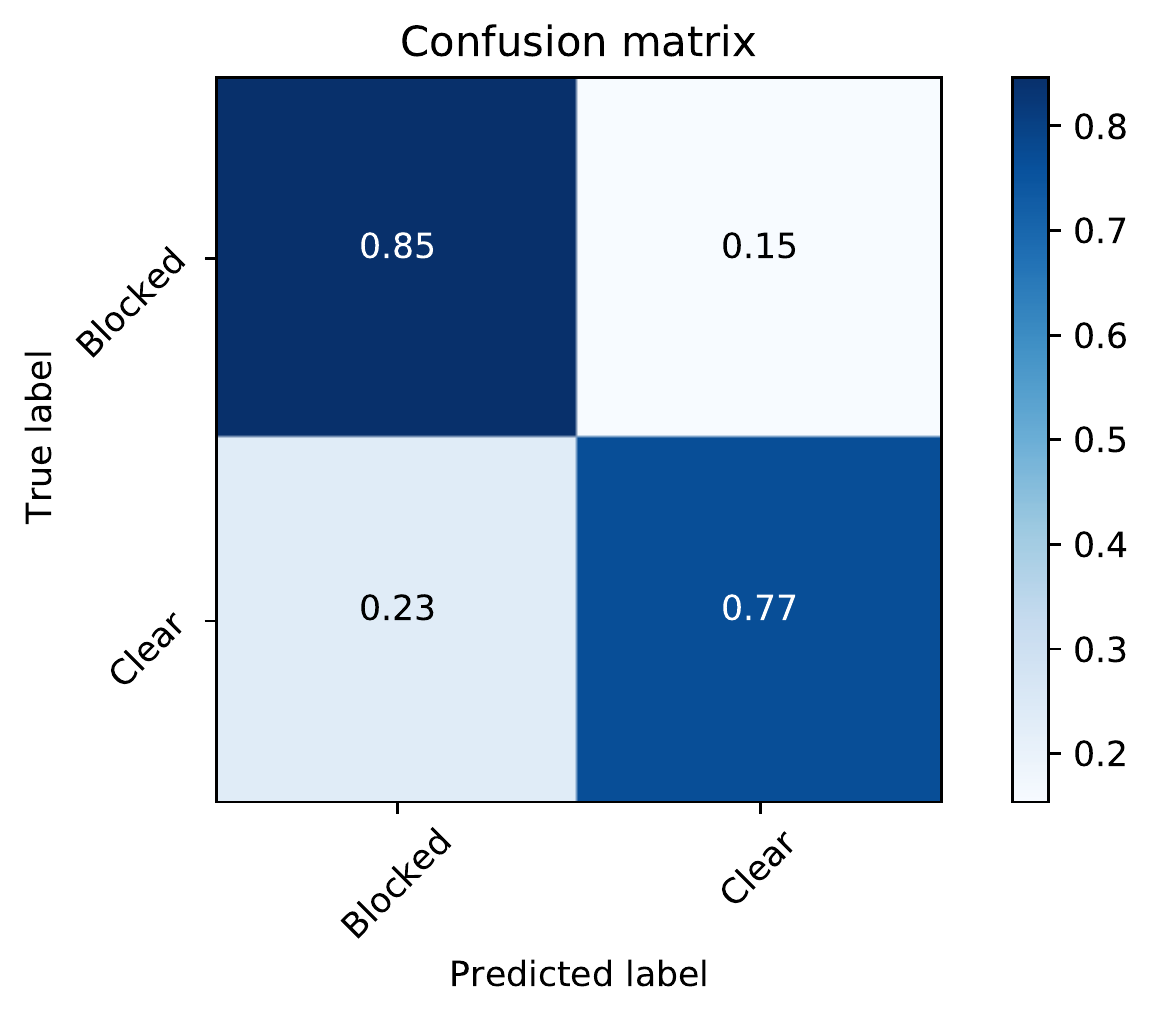}}
\subfigure[VGG16]{\includegraphics[scale=0.35]{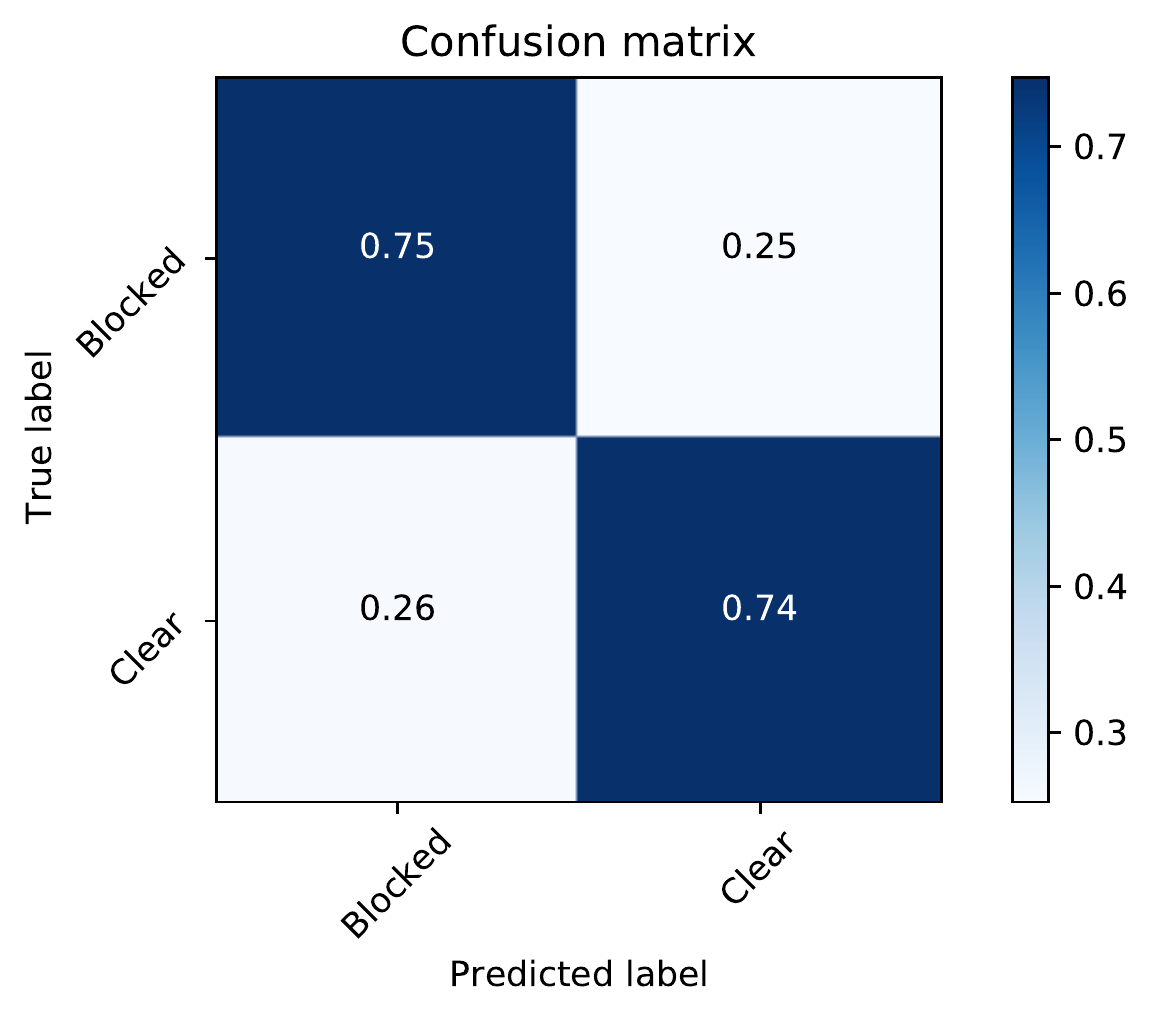}}
\subfigure[EfficientNetB3]{\includegraphics[scale=0.35]{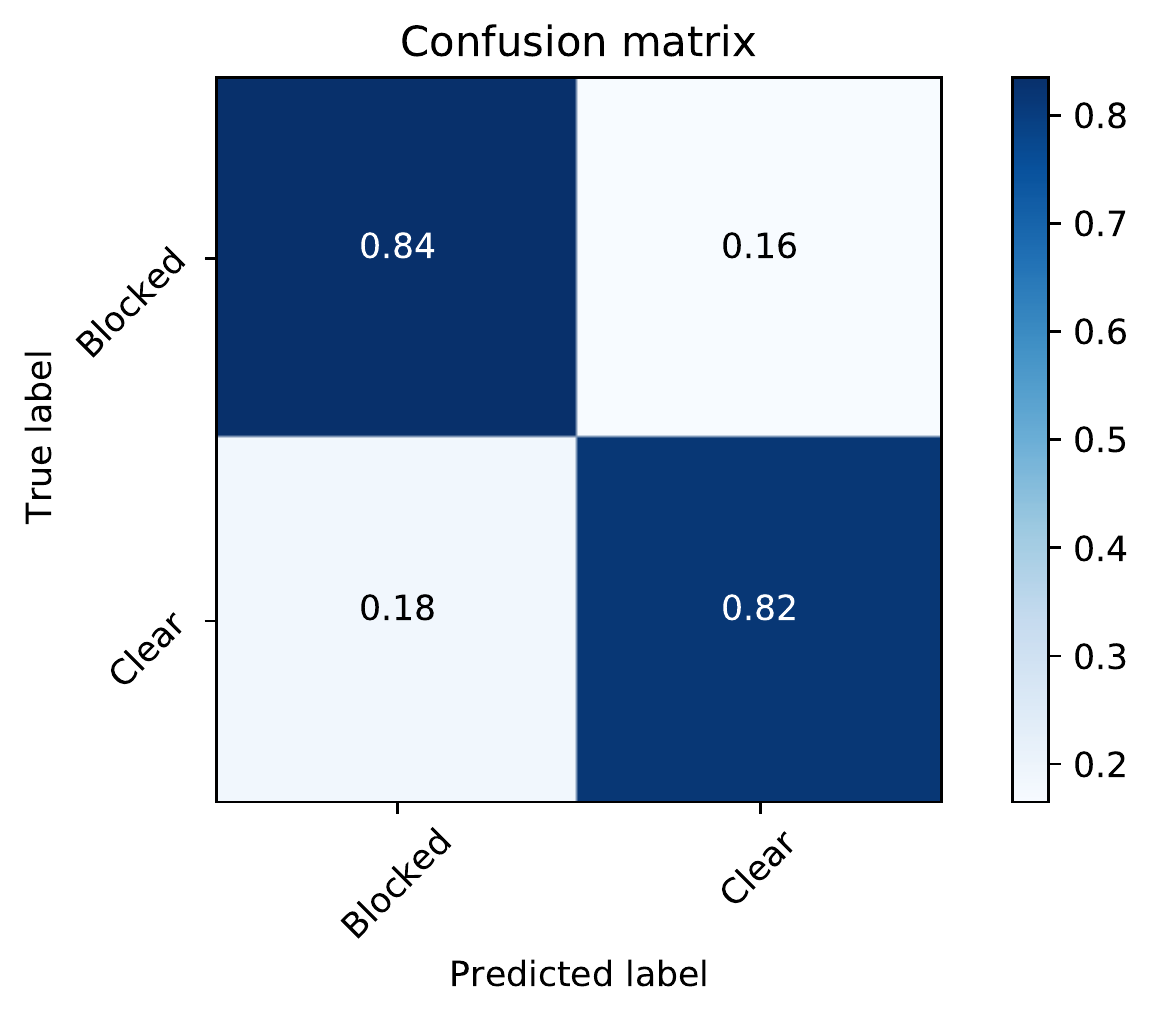}}\\
\subfigure[NASNet]{\includegraphics[scale=0.35]{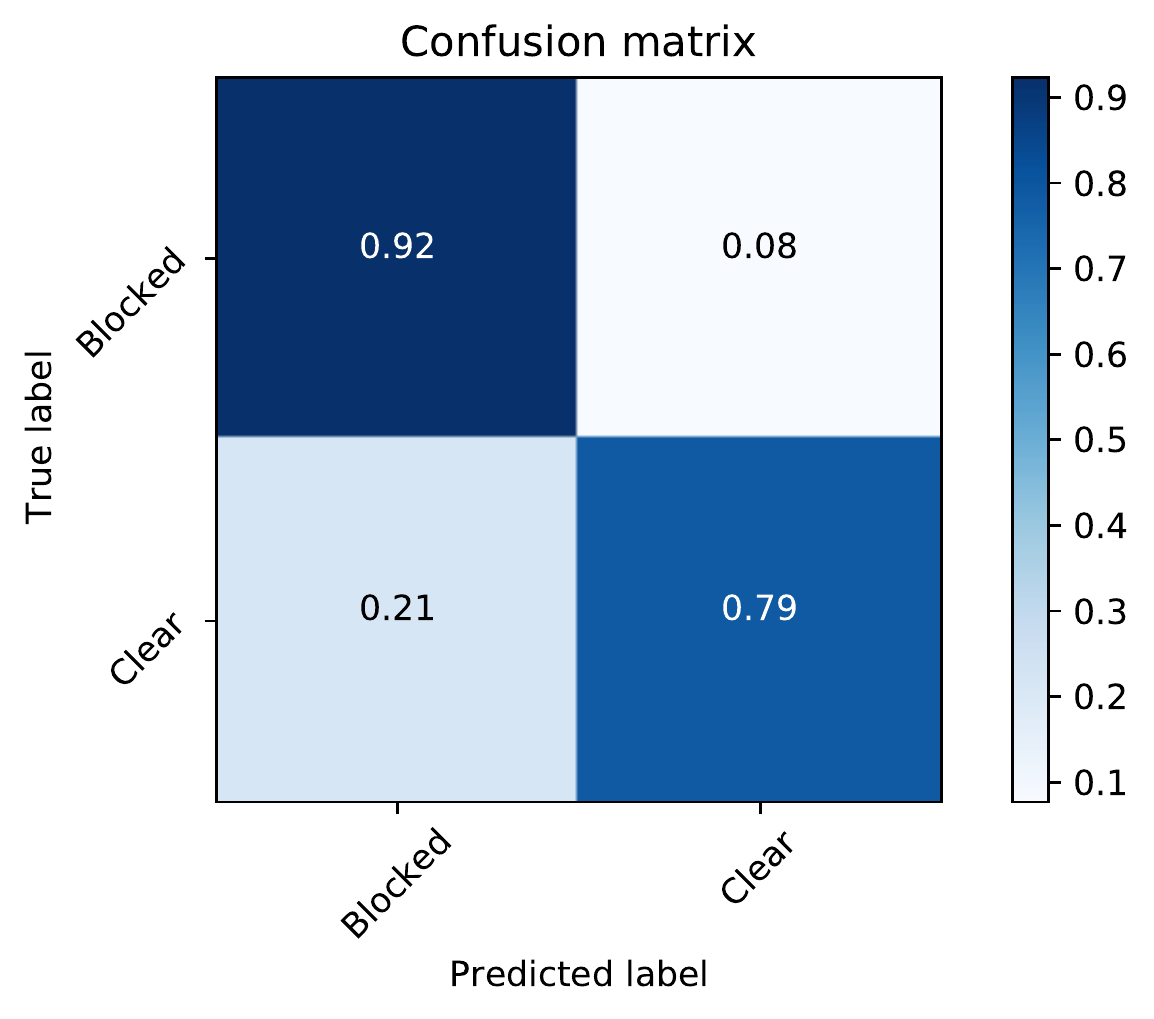}}
\caption{Confusion Matrices of Implemented CNN Models for Blockage Detection.}
    \label{fig:CM_Blockage}
\end{figure*}

\section{Experimental Setup and Evaluation Measures} \label{protocols}

Experiments were planned to investigate the performance of existing CNN models for binary classification of ICOB. Pre-trained CNN models with ImageNet weights were used for this investigation and implemented using Keras with Tensorflow at backend. Images of dimension $224 \times  224 \times  3$ were used as input to model and average pooling technique was used. Data augmentations techniques including samplewise standard deviation normalization, horizontal flip, vertical flip, rotation, width sift and height shift were used in the simulations for improved performance. All the models were tuned with dropout of 0.2, ReLU activation and batch normalization with SoftMax at fully connected layer. Stochastic Gradient Descent (SGD) optimizer with constant learning rate of 0.01 and categorical entropy loss was used. Each model was trained for 30 epochs with dataset divided into train, validate and test (60\%, 20\%, 20\%). All the simulations were performed using Nvidia GeForce RTX 2060 Graphical Processing Unit (GPU) with 6GB memory and 14 Gbps memory speed. Models were trained at full precision using Floating Point (FP-32) optimization. 

Performance of the models was measured in terms of their test accuracy, test loss, precision score, recall score, F1 score, Jaccard-Index, and processing times. In addition, confusion matrices were plotted to assess the Type I and Type II errors. Type I (False Positive (FP)) and Type II (False Negative (FN)) errors \cite{banerjee2009hypothesis} are commonly used terms in machine learning and main goal of model is to minimize one of these two errors, depending on context that which error is more critical in the given task. By definition, a Type I error is concluding the existence of a relationship while in fact it does not exist (e.g., classifying an image as ``blocked" while there is no blockage). Similarly, a Type II error is the rejection of the existence of relationship while in fact it exists (e.g., classifying an image as ``clear" while there is blockage). For the given culvert blockage context, Type II error is more critical to be minimized in comparison to Type I error because having notified as blocked while there is no blockage is tolerable in comparison to having notified as clear while there is blockage. Type II error will result in damages because it may be very late for response team to clear the blockage before diversion of flow.

\begin{table*}
    \caption{Model Processing Times for Three Different Size Images.}
    \label{tab:processing}
    \centering
    \begin{tabular}{ccccc}
    \toprule
    ~ & \multirow{2}{*}{\textbf{Model Processing Time (sec)}} & \multicolumn{3}{c}{\textbf{Total Execution Time (sec)}}\\
    ~ & ~ &   Image 1 & Image 2 & Image 3\\
    \toprule
DarkNet53	& 0.05		& 0.12	& 0.2	& 0.35\\
DenseNet121	& 0.09		& 0.17	& 0.24	& 0.39\\
InceptionResNetV2	& 0.14		& 0.21	& 0.29	& 0.44\\
InceptionV3	& 0.09		& 0.17	& 0.24	& 0.39\\
MobileNet	& 0.06		& 0.13	& 0.21	& 0.36\\
ResNet152	& 0.13		& 0.20	& 0.28	& 0.43\\
ResNet50	& 0.08		& 0.15	& 0.23	& 0.38\\
VGG16	& 0.08		& 0.15	& 0.23	& 0.38 \\
EfficientNetB3		& 0.09	& 0.16	& 0.24	& 0.39\\
NASNetLarge	& 0.15		& 0.22	& 0.30	& 0.45 \\
\bottomrule
    \end{tabular}
\end{table*}

\section{Results and Discussions}

Implemented CNN models were evaluated as per defined measures in Section \ref{protocols} and results were compared. Table \ref{tab:performance} presents the empirical results of all implemented models when evaluated for test dataset in terms of accuracy, loss, precision, recall, F1 score and Jaccard-Index. From the results, NASNet was reported as the best among all others with F1 score of 0.85. EfficientNetB3 and InceptionResNetV2 were reported as the second best with relatively same performance (F1 score of 0.83). DarkNet preformed worst with the F1 score of 0.71. 

Figure \ref{fig:CM_Blockage} shows the confusion matrices for the implemented CNN model to observe the Type I and Type II errors. From the figures, it can be observed that NASNet performed best in terms of lowest Type II error of only 8\%, however, Type I error was reported 21\%. On the other hand, EfficientNetB3 was reported with almost similar Type I and Type II errors (14\% and 16\%). From the FP instances, it was observed that for the cases where there are more than two openings and only one opening was blocked, algorithm classified it as clear. This insight lead to a suggestion in change of labelling criteria. A better approach could be to label image as blocked if half or more than half of openings are blocked otherwise label it as clear. Furthermore, if there is no debris material present in the image and occlusion is due to some foreground object not similar to debris in visual appearance, image should be labelled as clear. From the FN instances, it was observed that for the cases where image contained contents with visual appearance similar to blockage material, image was classified as blocked. This indicated the existence of background clutter/noise problem for this investigation.

\begin{comment}
\begin{figure*}
    \centering
    \includegraphics[scale=0.68]{Figures/FPFN.pdf}
    \caption{False Positive (First Row) and False Negative (Second Row) Instances.}
    \label{fig:Real_Samples}
\end{figure*}
\end{comment}

Implemented CNN models were also compared for their processing times to investigate the relative response times. Purpose of these analysis was to investigate the hardware implementability of proposed models for real-world applications. Model inference time and image processing time were calculated as two measures to compare the models. Three different size images were used; image 1 of $2048\times 1536$, image 2 of $3264\times 2448$ and image 3 of $4032\times 3024$. From the Table \ref{tab:processing}, it can be observed that MobileNet and DarkNet53 were fastest among others; however, were least accurate in this case. NASNet model was the slowest but most accurate in performance. As a trade-off, EfficientNetB3 model was relatively fast with accuracy towards higher end and recommended as a suitable choice to implement for on-board processing. It is important to mention that reported processing times are for relative comparison between models and not the actual measure of cutting edge hardware performance. However, given the availability of efficient computing hardware such Nvidia Jetson TX2 \cite{cui2019real} and Nvidia Jetson Nano \cite{basulto2020performance}, it is highly probable to implement any of the implemented models for real-world applications (e.g., pedestrian detection \cite{barthelemy2019edge}, wildlife tracking \cite{arshad2020my}).  

\begin{comment}
\begin{table*}[]
    \caption{Caption}
    \label{tab:my_label}
    \centering
    \begin{tabular}{cccccc}
    \toprule
    ~ & \multirow{2}{*}{\textbf{Model Processing Time (sec)}} & \multirow{2}{*}{\textbf{Model Load Time (sec)}} & \multicolumn{3}{c}{\textbf{Total Execution Time (sec)}}\\
    ~ & ~ & ~ &  Image 1 & Image 2 & Image 3\\
    \toprule
DarkNet53	& 0.05	& 13.3	& 0.12	& 0.2	& 0.35\\
DenseNet121	& 0.09	& 5.44	& 0.17	& 0.24	& 0.39\\
InceptionResNetV2	& 0.14	& 9.12	& 0.21	& 0.29	& 0.44\\
InceptionV3	& 0.09	& 5.75	& 0.17	& 0.24	& 0.39\\
MobileNet	& 0.06	& 2.90	& 0.13	& 0.21	& 0.36\\
ResNet152	& 0.13	& 19	& 0.20	& 0.28	& 0.43\\
ResNet50	& 0.08	& 4.16	& 0.15	& 0.23	& 0.38\\
VGG16	& 0.08	& 2.88	& 0.15	& 0.23	& 0.38 \\
EfficientNetB3	& 0.09	& 4.67	& 0.17	& 0.24	& 0.39\\
NASNetLarge	& 0.15	& 36.4	& 0.22	& 0.30	& 0.45 \\
\bottomrule
    \end{tabular}
\end{table*}

\begin{table*}
    \caption{Caption}
    \label{tab:my_label}
    \centering
    \begin{tabular}{cccc}
    \toprule
   ~ & \textbf{Image 1} & \textbf{Image 2} & \textbf{Image 3} \\
   ~ & $2048\times 1536$ & $3264\times 2448$ & $4032\times 3024$\\
    \toprule
Image Read Time (sec)	& 0.06	& 0.13	& 0.27\\
Image Resizing Time (sec)	& 0.01	& 0.02	& 0.03\\
Total Pre-Processing Time (sec)	& 0.07	& 0.15	& 0.30\\
\bottomrule
    \end{tabular}
\end{table*}
\end{comment}

\section{Conclusion and Future Directions}
Idea of using visual analytic for the culvert blockage analysis has been successfully pitched by implementing existing CNN models for culvert blockage classification. Images of Culvert Openings and Blockage (ICOB) dataset has been developed with diversity of clear and blocked culvert instances for training the CNN models. From the analysis, it has been observed that NASNet model performed best among all in terms of classification performance, however, was the slowest in relative comparison of processing times. Based on the classification performance and processing times, EfficientNetB3 was recommended model to be deployed for real-world application. From the FP and FN instances, background noise and oversimplified labelling criteria were found potential factors for degraded performance. A visual attention based algorithm and/or detection-classification pipeline are the concepts that can be implemented to address the background noise problem. Furthermore, enhancement of dataset by injecting scaled physical model and computer generated synthetic images are potential future directions. 

\section*{Acknowledgment}
I would like to thank WCC for providing resources and support to carryout this study. Furthermore, I would like to thank University of Wollongong (UOW) and Higher Education Commission (HEC) of Pakistan for funding my PhD studies.

\balance
\bibliographystyle{ACM-Reference-Format}
\bibliography{References}

%%% -*-BibTeX-*-
%%% Do NOT edit. File created by BibTeX with style
%%% ACM-Reference-Format-Journals [18-Jan-2012].

\begin{thebibliography}{32}

%%% ====================================================================
%%% NOTE TO THE USER: you can override these defaults by providing
%%% customized versions of any of these macros before the \bibliography
%%% command.  Each of them MUST provide its own final punctuation,
%%% except for \shownote{}, \showDOI{}, and \showURL{}.  The latter two
%%% do not use final punctuation, in order to avoid confusing it with
%%% the Web address.
%%%
%%% To suppress output of a particular field, define its macro to expand
%%% to an empty string, or better, \unskip, like this:
%%%
%%% \newcommand{\showDOI}[1]{\unskip}   % LaTeX syntax
%%%
%%% \def \showDOI #1{\unskip}           % plain TeX syntax
%%%
%%% ====================================================================

\ifx \showCODEN    \undefined \def \showCODEN     #1{\unskip}     \fi
\ifx \showDOI      \undefined \def \showDOI       #1{#1}\fi
\ifx \showISBNx    \undefined \def \showISBNx     #1{\unskip}     \fi
\ifx \showISBNxiii \undefined \def \showISBNxiii  #1{\unskip}     \fi
\ifx \showISSN     \undefined \def \showISSN      #1{\unskip}     \fi
\ifx \showLCCN     \undefined \def \showLCCN      #1{\unskip}     \fi
\ifx \shownote     \undefined \def \shownote      #1{#1}          \fi
\ifx \showarticletitle \undefined \def \showarticletitle #1{#1}   \fi
\ifx \showURL      \undefined \def \showURL       {\relax}        \fi
% The following commands are used for tagged output and should be
% invisible to TeX
\providecommand\bibfield[2]{#2}
\providecommand\bibinfo[2]{#2}
\providecommand\natexlab[1]{#1}
\providecommand\showeprint[2][]{arXiv:#2}

\bibitem[\protect\citeauthoryear{Arshad, Barthelemy, Pilton, and Perez}{Arshad
  et~al\mbox{.}}{2020}]%
        {arshad2020my}
\bibfield{author}{\bibinfo{person}{Bilal Arshad}, \bibinfo{person}{Johan
  Barthelemy}, \bibinfo{person}{Elliott Pilton}, {and} \bibinfo{person}{Pascal
  Perez}.} \bibinfo{year}{2020}\natexlab{}.
\newblock \showarticletitle{Where is my Deer?-Wildlife Tracking And Counting
  via Edge Computing And Deep Learning}. In \bibinfo{booktitle}{\emph{2020 IEEE
  Sensors}}. \bibinfo{publisher}{IEEE}, \bibinfo{address}{Rotterdam,
  Netherlands}, \bibinfo{pages}{1--4}.
\newblock


\bibitem[\protect\citeauthoryear{Arshad, Ogie, Barthelemy, Pradhan, Verstaevel,
  and Perez}{Arshad et~al\mbox{.}}{2019}]%
        {arshad2019computer}
\bibfield{author}{\bibinfo{person}{Bilal Arshad}, \bibinfo{person}{Robert
  Ogie}, \bibinfo{person}{Johan Barthelemy}, \bibinfo{person}{Biswajeet
  Pradhan}, \bibinfo{person}{Nicolas Verstaevel}, {and} \bibinfo{person}{Pascal
  Perez}.} \bibinfo{year}{2019}\natexlab{}.
\newblock \showarticletitle{Computer Vision and IoT-Based Sensors in Flood
  Monitoring and Mapping: A Systematic Review}.
\newblock \bibinfo{journal}{\emph{Sensors}} \bibinfo{volume}{19},
  \bibinfo{number}{22} (\bibinfo{year}{2019}), \bibinfo{pages}{5012}.
\newblock


\bibitem[\protect\citeauthoryear{Ball, Babister, Nathan, Weinmann, Weeks,
  Retallick, and Testoni}{Ball et~al\mbox{.}}{2016}]%
        {ball2016australian}
\bibfield{author}{\bibinfo{person}{JE Ball}, \bibinfo{person}{MK Babister},
  \bibinfo{person}{R Nathan}, \bibinfo{person}{PE Weinmann}, \bibinfo{person}{W
  Weeks}, \bibinfo{person}{M Retallick}, {and} \bibinfo{person}{I Testoni}.}
  \bibinfo{year}{2016}\natexlab{}.
\newblock \bibinfo{title}{Australian Rainfall and Runoff-A guide to flood
  estimation}.
\newblock \bibinfo{howpublished}{Commonwealth of Australia}.
\newblock


\bibitem[\protect\citeauthoryear{Banerjee, Chitnis, Jadhav, Bhawalkar, and
  Chaudhury}{Banerjee et~al\mbox{.}}{2009}]%
        {banerjee2009hypothesis}
\bibfield{author}{\bibinfo{person}{Amitav Banerjee}, \bibinfo{person}{UB
  Chitnis}, \bibinfo{person}{SL Jadhav}, \bibinfo{person}{JS Bhawalkar}, {and}
  \bibinfo{person}{S Chaudhury}.} \bibinfo{year}{2009}\natexlab{}.
\newblock \showarticletitle{Hypothesis testing, type I and type II errors}.
\newblock \bibinfo{journal}{\emph{Industrial psychiatry journal}}
  \bibinfo{volume}{18}, \bibinfo{number}{2} (\bibinfo{year}{2009}),
  \bibinfo{pages}{127}.
\newblock


\bibitem[\protect\citeauthoryear{Barth{\'e}lemy, Verstaevel, Forehead, and
  Perez}{Barth{\'e}lemy et~al\mbox{.}}{2019}]%
        {barthelemy2019edge}
\bibfield{author}{\bibinfo{person}{Johan Barth{\'e}lemy},
  \bibinfo{person}{Nicolas Verstaevel}, \bibinfo{person}{Hugh Forehead}, {and}
  \bibinfo{person}{Pascal Perez}.} \bibinfo{year}{2019}\natexlab{}.
\newblock \showarticletitle{Edge-computing video analytics for real-time
  traffic monitoring in a smart city}.
\newblock \bibinfo{journal}{\emph{Sensors}} \bibinfo{volume}{19},
  \bibinfo{number}{9} (\bibinfo{year}{2019}), \bibinfo{pages}{2048}.
\newblock


\bibitem[\protect\citeauthoryear{Barthelmess and Rigby}{Barthelmess and
  Rigby}{2011}]%
        {BarthelmessMechanism}
\bibfield{author}{\bibinfo{person}{AJ Barthelmess} {and} \bibinfo{person}{EH
  Rigby}.} \bibinfo{year}{2011}\natexlab{}.
\newblock \showarticletitle{Culvert Blockage Mechanisms and their Impact on
  Flood Behaviour}. In \bibinfo{booktitle}{\emph{Proceedings of the 34th World
  Congress of the International Association for Hydro- Environment Research and
  Engineering}}. \bibinfo{publisher}{Engineers Australia},
  \bibinfo{address}{Barton, ACT}, \bibinfo{pages}{380--387}.
\newblock


\bibitem[\protect\citeauthoryear{Basulto-Lantsova, Padilla-Medina, Perez-Pinal,
  and Barranco-Gutierrez}{Basulto-Lantsova et~al\mbox{.}}{2020}]%
        {basulto2020performance}
\bibfield{author}{\bibinfo{person}{Artiom Basulto-Lantsova},
  \bibinfo{person}{Jose~A Padilla-Medina}, \bibinfo{person}{Francisco~J
  Perez-Pinal}, {and} \bibinfo{person}{Alejandro~I Barranco-Gutierrez}.}
  \bibinfo{year}{2020}\natexlab{}.
\newblock \showarticletitle{Performance comparative of OpenCV Template Matching
  method on Jetson TX2 and Jetson Nano developer kits}. In
  \bibinfo{booktitle}{\emph{2020 10th Annual Computing and Communication
  Workshop and Conference (CCWC)}}. \bibinfo{publisher}{IEEE},
  \bibinfo{address}{Las Vegas, NV, USA}, \bibinfo{pages}{0812--0816}.
\newblock


\bibitem[\protect\citeauthoryear{Blanc}{Blanc}{2013}]%
        {blanc2013analysis}
\bibfield{author}{\bibinfo{person}{Janice Blanc}.}
  \bibinfo{year}{2013}\natexlab{}.
\newblock \emph{\bibinfo{title}{An analysis of the impact of trash screen
  design on debris related blockage at culvert inlets}}.
\newblock \bibinfo{thesistype}{Ph.D. Dissertation}. \bibinfo{school}{School of
  the Built Environment, Heriot-Watt University}.
\newblock


\bibitem[\protect\citeauthoryear{Cui and Dahnoun}{Cui and Dahnoun}{2019}]%
        {cui2019real}
\bibfield{author}{\bibinfo{person}{Han Cui} {and} \bibinfo{person}{Naim
  Dahnoun}.} \bibinfo{year}{2019}\natexlab{}.
\newblock \showarticletitle{Real-Time Stereo Vision Implementation on Nvidia
  Jetson TX2}. In \bibinfo{booktitle}{\emph{2019 8th Mediterranean Conference
  on Embedded Computing (MECO)}}. \bibinfo{publisher}{IEEE},
  \bibinfo{address}{Budva, Montenegro}, \bibinfo{pages}{1--5}.
\newblock


\bibitem[\protect\citeauthoryear{Davis}{Davis}{2001}]%
        {davis2001analysis}
\bibfield{author}{\bibinfo{person}{AJ Davis}.} \bibinfo{year}{2001}\natexlab{}.
\newblock \bibinfo{title}{An analysis of the effects of debris caught at
  various points of major catchments during Wollongong’s August 1998 storm
  event}.
\newblock \bibinfo{howpublished}{Bachelor of Engineering Thesis, University of
  Wollongong}.
\newblock


\bibitem[\protect\citeauthoryear{French and Jones}{French and Jones}{2015}]%
        {french2015culvert}
\bibfield{author}{\bibinfo{person}{Robert French} {and}
  \bibinfo{person}{Malcolm Jones}.} \bibinfo{year}{2015}\natexlab{}.
\newblock \showarticletitle{Culvert blockages in two Australian flood events
  and implications for design}.
\newblock \bibinfo{journal}{\emph{Australasian Journal of Water Resources}}
  \bibinfo{volume}{19}, \bibinfo{number}{2} (\bibinfo{year}{2015}),
  \bibinfo{pages}{134--142}.
\newblock


\bibitem[\protect\citeauthoryear{French and Jones}{French and Jones}{2018}]%
        {french2018design}
\bibfield{author}{\bibinfo{person}{Robert French} {and}
  \bibinfo{person}{Malcolm Jones}.} \bibinfo{year}{2018}\natexlab{}.
\newblock \showarticletitle{Design for culvert blockage: the ARR 2016
  guidelines}.
\newblock \bibinfo{journal}{\emph{Australasian Journal of Water Resources}}
  \bibinfo{volume}{22}, \bibinfo{number}{1} (\bibinfo{year}{2018}),
  \bibinfo{pages}{84--87}.
\newblock


\bibitem[\protect\citeauthoryear{French, Rigby, and Barthelmess}{French
  et~al\mbox{.}}{2012}]%
        {french2012non}
\bibfield{author}{\bibinfo{person}{R French}, \bibinfo{person}{EH Rigby}, {and}
  \bibinfo{person}{AJ Barthelmess}.} \bibinfo{year}{2012}\natexlab{}.
\newblock \showarticletitle{The non-impact of debris blockages on the August
  1998 Wollongong flooding}.
\newblock \bibinfo{journal}{\emph{Australasian Journal of Water Resources}}
  \bibinfo{volume}{15}, \bibinfo{number}{2} (\bibinfo{year}{2012}),
  \bibinfo{pages}{161--169}.
\newblock


\bibitem[\protect\citeauthoryear{He, Zhang, Ren, and Sun}{He
  et~al\mbox{.}}{2016}]%
        {he2016deep}
\bibfield{author}{\bibinfo{person}{Kaiming He}, \bibinfo{person}{Xiangyu
  Zhang}, \bibinfo{person}{Shaoqing Ren}, {and} \bibinfo{person}{Jian Sun}.}
  \bibinfo{year}{2016}\natexlab{}.
\newblock \showarticletitle{Deep residual learning for image recognition}. In
  \bibinfo{booktitle}{\emph{Proceedings of the IEEE conference on computer
  vision and pattern recognition}}. \bibinfo{publisher}{IEEE},
  \bibinfo{address}{Las Vegas, NV, USA}, \bibinfo{pages}{770--778}.
\newblock


\bibitem[\protect\citeauthoryear{Howard, Zhu, Chen, Kalenichenko, Wang, Weyand,
  Andreetto, and Adam}{Howard et~al\mbox{.}}{2017}]%
        {howard2017mobilenets}
\bibfield{author}{\bibinfo{person}{Andrew~G Howard}, \bibinfo{person}{Menglong
  Zhu}, \bibinfo{person}{Bo Chen}, \bibinfo{person}{Dmitry Kalenichenko},
  \bibinfo{person}{Weijun Wang}, \bibinfo{person}{Tobias Weyand},
  \bibinfo{person}{Marco Andreetto}, {and} \bibinfo{person}{Hartwig Adam}.}
  \bibinfo{year}{2017}\natexlab{}.
\newblock \showarticletitle{Mobilenets: Efficient convolutional neural networks
  for mobile vision applications}.
\newblock \bibinfo{journal}{\emph{arXiv preprint arXiv:1704.04861}}
  (\bibinfo{year}{2017}).
\newblock


\bibitem[\protect\citeauthoryear{Huang, Liu, Van Der~Maaten, and
  Weinberger}{Huang et~al\mbox{.}}{2017}]%
        {huang2017densely}
\bibfield{author}{\bibinfo{person}{Gao Huang}, \bibinfo{person}{Zhuang Liu},
  \bibinfo{person}{Laurens Van Der~Maaten}, {and} \bibinfo{person}{Kilian~Q
  Weinberger}.} \bibinfo{year}{2017}\natexlab{}.
\newblock \showarticletitle{Densely connected convolutional networks}. In
  \bibinfo{booktitle}{\emph{Proceedings of the IEEE conference on computer
  vision and pattern recognition}}. \bibinfo{publisher}{IEEE},
  \bibinfo{address}{Honolulu, Hawaii}, \bibinfo{pages}{4700--4708}.
\newblock


\bibitem[\protect\citeauthoryear{Iqbal, Perez, Li, and Barthelemy}{Iqbal
  et~al\mbox{.}}{2021}]%
        {iqbalcomputer}
\bibfield{author}{\bibinfo{person}{Umair Iqbal}, \bibinfo{person}{Pascal
  Perez}, \bibinfo{person}{Wanqing Li}, {and} \bibinfo{person}{Johan
  Barthelemy}.} \bibinfo{year}{2021}\natexlab{}.
\newblock \showarticletitle{How Computer Vision can Facilitate Flood
  Management: A Systematic Review}.
\newblock \bibinfo{journal}{\emph{International Journal of Disaster Risk
  Reduction}}  \bibinfo{volume}{53} (\bibinfo{year}{2021}),
  \bibinfo{pages}{102030}.
\newblock


\bibitem[\protect\citeauthoryear{Jones, Weeks, and Babister}{Jones
  et~al\mbox{.}}{2016}]%
        {CondouitBlockagePolicy}
\bibfield{author}{\bibinfo{person}{Rhys~Hardwick Jones},
  \bibinfo{person}{William Weeks}, {and} \bibinfo{person}{Mark Babister}.}
  \bibinfo{year}{2016}\natexlab{}.
\newblock \bibinfo{booktitle}{\emph{Review of Conduit Blockage Policy Summary
  Report}}.
\newblock \bibinfo{publisher}{WMA Water}, \bibinfo{address}{160 Clarence Street
  Sydney, NSW, 2000}.
\newblock


\bibitem[\protect\citeauthoryear{Ollett, Syme, and Ryan}{Ollett
  et~al\mbox{.}}{2017}]%
        {ollett2017australian}
\bibfield{author}{\bibinfo{person}{Paul Ollett}, \bibinfo{person}{Bill Syme},
  {and} \bibinfo{person}{Phil Ryan}.} \bibinfo{year}{2017}\natexlab{}.
\newblock \showarticletitle{Australian Rainfall and Runoff guidance on blockage
  of hydraulic structures: numerical implementation and three case studies}.
\newblock \bibinfo{journal}{\emph{Journal of Hydrology (New Zealand)}}
  \bibinfo{volume}{56}, \bibinfo{number}{2} (\bibinfo{year}{2017}),
  \bibinfo{pages}{109--122}.
\newblock


\bibitem[\protect\citeauthoryear{Redmon and Farhadi}{Redmon and
  Farhadi}{2018}]%
        {redmon2018yolov3}
\bibfield{author}{\bibinfo{person}{Joseph Redmon} {and} \bibinfo{person}{Ali
  Farhadi}.} \bibinfo{year}{2018}\natexlab{}.
\newblock \showarticletitle{Yolov3: An incremental improvement}.
\newblock \bibinfo{journal}{\emph{arXiv preprint arXiv:1804.02767}}
  (\bibinfo{year}{2018}).
\newblock


\bibitem[\protect\citeauthoryear{Rigby and Silveri}{Rigby and Silveri}{2001}]%
        {rigby2001impact}
\bibfield{author}{\bibinfo{person}{EH Rigby} {and} \bibinfo{person}{P
  Silveri}.} \bibinfo{year}{2001}\natexlab{}.
\newblock \showarticletitle{The impact of blockages on flood behaviour in the
  Wollongong storm of August 1998}. In \bibinfo{booktitle}{\emph{6th Conference
  on Hydraulics in Civil Engineering: The State of Hydraulics}}.
  \bibinfo{publisher}{Engineers Australia}, \bibinfo{address}{Barton, ACT},
  \bibinfo{pages}{107--115}.
\newblock


\bibitem[\protect\citeauthoryear{Rigby and Silveri}{Rigby and Silveri}{2002}]%
        {rigby2002causes}
\bibfield{author}{\bibinfo{person}{EH Rigby} {and} \bibinfo{person}{P
  Silveri}.} \bibinfo{year}{2002}\natexlab{}.
\newblock \showarticletitle{Causes and effects of culvert blockage during large
  storms}. In \bibinfo{booktitle}{\emph{Ninth International Conference on Urban
  Drainage (9ICUD)}}. \bibinfo{publisher}{Engineers Australia},
  \bibinfo{address}{Lloyd Center Doubletree Hotel, Portland, Oregon, United
  States}, \bibinfo{pages}{1--16}.
\newblock


\bibitem[\protect\citeauthoryear{Roso, Boyd, Rigby, and VanDrie}{Roso
  et~al\mbox{.}}{2004}]%
        {roso2004prediction}
\bibfield{author}{\bibinfo{person}{Steven Roso}, \bibinfo{person}{Michael
  Boyd}, \bibinfo{person}{E Rigby}, {and} \bibinfo{person}{Rudy VanDrie}.}
  \bibinfo{year}{2004}\natexlab{}.
\newblock \showarticletitle{Prediction of increased flooding in urban
  catchments due to debris blockage and flow diversions}. In
  \bibinfo{booktitle}{\emph{Proceedings of the 5th International Conference on
  Sustainable Techniques and Strategies in Urban Water Management (NOVATECH)}}.
  \bibinfo{publisher}{Water Science and Technology}, \bibinfo{address}{Lyon,
  France}, \bibinfo{pages}{8--13}.
\newblock


\bibitem[\protect\citeauthoryear{Simonyan and Zisserman}{Simonyan and
  Zisserman}{2014}]%
        {simonyan2014very}
\bibfield{author}{\bibinfo{person}{Karen Simonyan} {and}
  \bibinfo{person}{Andrew Zisserman}.} \bibinfo{year}{2014}\natexlab{}.
\newblock \showarticletitle{Very deep convolutional networks for large-scale
  image recognition}.
\newblock \bibinfo{journal}{\emph{arXiv preprint arXiv:1409.1556}}
  (\bibinfo{year}{2014}).
\newblock


\bibitem[\protect\citeauthoryear{Szegedy, Ioffe, Vanhoucke, and Alemi}{Szegedy
  et~al\mbox{.}}{2016a}]%
        {szegedy2016inception}
\bibfield{author}{\bibinfo{person}{Christian Szegedy}, \bibinfo{person}{Sergey
  Ioffe}, \bibinfo{person}{Vincent Vanhoucke}, {and} \bibinfo{person}{Alex
  Alemi}.} \bibinfo{year}{2016}\natexlab{a}.
\newblock \showarticletitle{Inception-v4, inception-resnet and the impact of
  residual connections on learning}.
\newblock \bibinfo{journal}{\emph{arXiv preprint arXiv:1602.07261}}
  (\bibinfo{year}{2016}).
\newblock


\bibitem[\protect\citeauthoryear{Szegedy, Vanhoucke, Ioffe, Shlens, and
  Wojna}{Szegedy et~al\mbox{.}}{2016b}]%
        {szegedy2016rethinking}
\bibfield{author}{\bibinfo{person}{Christian Szegedy}, \bibinfo{person}{Vincent
  Vanhoucke}, \bibinfo{person}{Sergey Ioffe}, \bibinfo{person}{Jon Shlens},
  {and} \bibinfo{person}{Zbigniew Wojna}.} \bibinfo{year}{2016}\natexlab{b}.
\newblock \showarticletitle{Rethinking the inception architecture for computer
  vision}. In \bibinfo{booktitle}{\emph{Proceedings of the IEEE conference on
  computer vision and pattern recognition}}. \bibinfo{publisher}{IEEE},
  \bibinfo{address}{Las Vegas, NV, USA}, \bibinfo{pages}{2818--2826}.
\newblock


\bibitem[\protect\citeauthoryear{Tan and Le}{Tan and Le}{2019}]%
        {tan2019efficientnet}
\bibfield{author}{\bibinfo{person}{Mingxing Tan} {and} \bibinfo{person}{Quoc~V
  Le}.} \bibinfo{year}{2019}\natexlab{}.
\newblock \showarticletitle{Efficientnet: Rethinking model scaling for
  convolutional neural networks}.
\newblock \bibinfo{journal}{\emph{arXiv preprint arXiv:1905.11946}}
  (\bibinfo{year}{2019}).
\newblock


\bibitem[\protect\citeauthoryear{Van~Drie, Boyd, and Rigby}{Van~Drie
  et~al\mbox{.}}{2001}]%
        {van2001modelling}
\bibfield{author}{\bibinfo{person}{R Van~Drie}, \bibinfo{person}{MJ Boyd},
  {and} \bibinfo{person}{EH Rigby}.} \bibinfo{year}{2001}\natexlab{}.
\newblock \showarticletitle{Modelling of hydraulic flood flows using WBNM2001}.
  In \bibinfo{booktitle}{\emph{6th Conference on Hydraulics in Civil
  Engineering}}. \bibinfo{publisher}{Institution of Engineers Australia},
  \bibinfo{address}{Hobart, Australia}, \bibinfo{pages}{523--531}.
\newblock


\bibitem[\protect\citeauthoryear{Wallerstein, Thorne, and Abt}{Wallerstein
  et~al\mbox{.}}{1996}]%
        {wallerstein1996debris}
\bibfield{author}{\bibinfo{person}{N Wallerstein}, \bibinfo{person}{Colin~R
  Thorne}, {and} \bibinfo{person}{S Abt}.} \bibinfo{year}{1996}\natexlab{}.
\newblock \bibinfo{booktitle}{\emph{Debris control at hydraulic structures,
  contract modification: management of woody debris in natural channels and at
  hydraulic structures}}.
\newblock \bibinfo{type}{{T}echnical {R}eport}.
  \bibinfo{institution}{Nottingham University (United Kingdom) Department of
  Geography}.
\newblock


\bibitem[\protect\citeauthoryear{WBM}{WBM}{2008}]%
        {wbm2008newcastle}
\bibfield{author}{\bibinfo{person}{BMT WBM}.} \bibinfo{year}{2008}\natexlab{}.
\newblock \showarticletitle{Newcastle Flash Flood 8 June 2007 (the Pasha Bulker
  Storm) Flood Data Compendium}.
\newblock \bibinfo{journal}{\emph{Prepared for Newcastle City Council, BMT WBM,
  Broadmeadow}} (\bibinfo{year}{2008}).
\newblock


\bibitem[\protect\citeauthoryear{Weeks, Witheridge, Rigby, Barthelmess, and
  O‘Loughlin}{Weeks et~al\mbox{.}}{2013}]%
        {ARR_Report}
\bibfield{author}{\bibinfo{person}{W Weeks}, \bibinfo{person}{G Witheridge},
  \bibinfo{person}{E Rigby}, \bibinfo{person}{A Barthelmess}, {and}
  \bibinfo{person}{G O‘Loughlin}.} \bibinfo{year}{2013}\natexlab{}.
\newblock \bibinfo{booktitle}{\emph{Project 11: Blockage of Hydraulic
  Structures}}.
\newblock \bibinfo{type}{{T}echnical {R}eport} P11/S2/021.
  \bibinfo{institution}{Engineers Australia, Water Engineering},
  \bibinfo{address}{11 National Circuit Barton ACT 2600}.
\newblock


\bibitem[\protect\citeauthoryear{Zoph, Vasudevan, Shlens, and Le}{Zoph
  et~al\mbox{.}}{2018}]%
        {zoph2018learning}
\bibfield{author}{\bibinfo{person}{Barret Zoph}, \bibinfo{person}{Vijay
  Vasudevan}, \bibinfo{person}{Jonathon Shlens}, {and} \bibinfo{person}{Quoc~V
  Le}.} \bibinfo{year}{2018}\natexlab{}.
\newblock \showarticletitle{Learning transferable architectures for scalable
  image recognition}. In \bibinfo{booktitle}{\emph{Proceedings of the IEEE
  conference on computer vision and pattern recognition}}.
  \bibinfo{publisher}{IEEE}, \bibinfo{address}{Salt Lake City, USA},
  \bibinfo{pages}{8697--8710}.
\newblock


\end{thebibliography}

\end{document}